\pdfoutput=1

\documentclass[11pt]{article}

\usepackage[final]{acl}

\usepackage{times}
\usepackage{latexsym}

\usepackage[T1]{fontenc}

\usepackage[utf8]{inputenc}

\usepackage{microtype}

\usepackage{inconsolata}

\usepackage{graphicx}

\usepackage{fdsymbol}

\usepackage{svg}
\title{\includegraphics[scale=0.027]{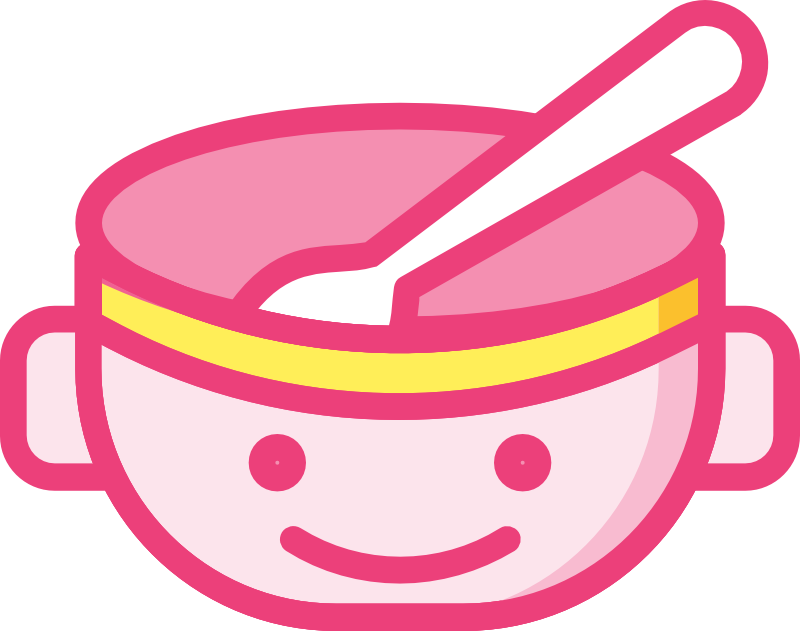}\hspace{1pt} \textsc{VectorEdits}:
 A Dataset and Benchmark \\for Instruction-Based Editing of Vector Graphics}

\author{
  Josef Kuchař\textsuperscript{$\spadesuit$} \textnormal{\and}
  Marek Kadlčík\textsuperscript{$\spadesuit$} \textnormal{\and}
  Michal Spiegel\textsuperscript{$\spadesuit\diamondsuit$} \textnormal{\and}
  \vspace{5pt}
  Michal Štefánik\textsuperscript{$\heartsuit\spadesuit$} \\
  \textsuperscript{$\spadesuit$}TransformersClub @
        Faculty of Informatics, Masaryk University\vspace{-2pt} \\
  \textsuperscript{$\diamondsuit$}Kempelen Institute of Intelligent Technologies\vspace{-2pt} \\
  \textsuperscript{$\heartsuit$}Language Technology, University of Helsinki \\
}

\begin{document}
\maketitle
\begin{abstract}

We introduce a large-scale dataset for instruction-guided vector image editing, consisting of over 270,000 pairs of SVG images paired with natural language edit instructions. Our dataset enables training and evaluation of models that modify vector graphics based on textual commands. We describe the data collection process, including image pairing via CLIP similarity and instruction generation with vision-language models. Initial experiments with state-of-the-art large language models reveal that current methods struggle to produce accurate and valid edits, underscoring the challenge of this task. To foster research in natural language-driven vector graphic generation and editing, we make our resources created within this work publicly available.\footnote{Our dataset can be found on \url{https://huggingface.co/datasets/mikronai/VectorEdits}}
\end{abstract}

\section{Introduction}

Vector graphics play a crucial role in modern digital content creation, enabling scalable, resolution-independent visual elements across web, print, and user interface design. Unlike raster images, vector graphics are composed of geometric primitives such as paths, curves, and shapes, making them highly editable and efficient for a wide range of design tasks.

In this work, we focus on the task of instruction-guided vector image editing—modifying a vector graphic based on a natural language instruction. This is a challenging problem that requires a model to combine multiple advanced capabilities: visual understanding to interpret the source image, spatial reasoning to identify and localize elements described in the instruction, and code generation to produce valid and semantically meaningful SVG edits.

Beyond its technical complexity, this task has meaningful practical implications. Progress in instruction-based SVG generation and editing could significantly lower the barrier to entry for digital creativity, enabling the free creation, customization, and sharing of vector-based digital art. Such systems could support both novice users and professionals in rapidly prototyping, adapting, and remixing designs through simple language commands.

\begin{figure}[t]
\centering
  \includegraphics[width=\columnwidth]{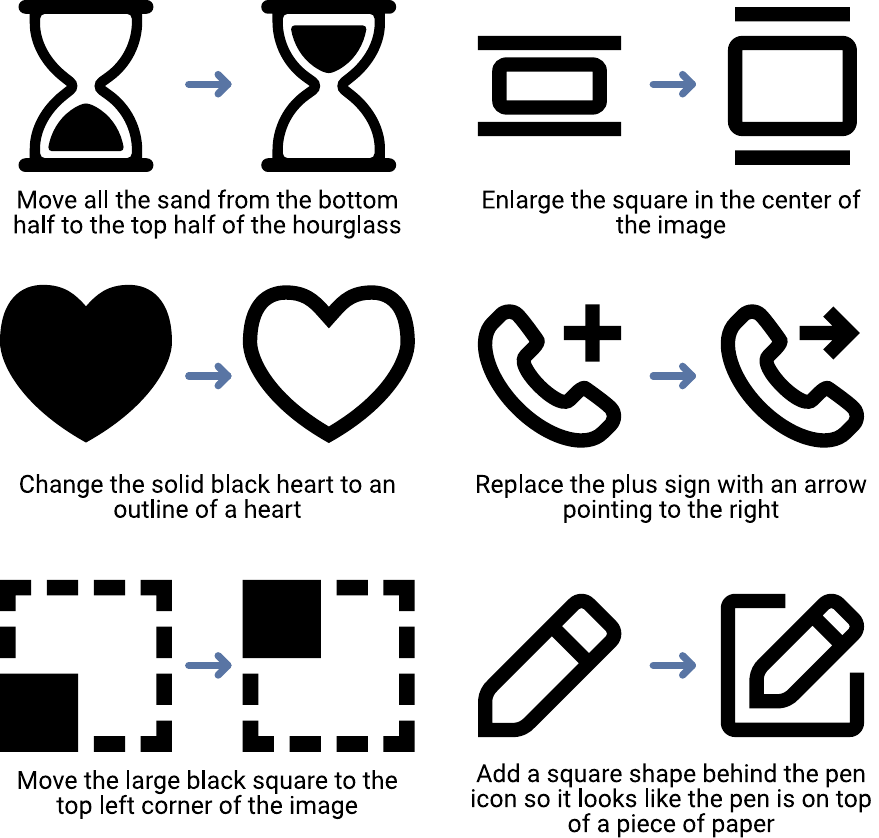}
  \caption{Examples of instruction-guided vector image editing pairs from our dataset.}
  \label{fig:example_pair}
\end{figure}

\section{Background}

Several recent efforts have explored vision-language tasks in the vector domain. Datasets like SVGBench~\cite{https://doi.org/10.48550/arxiv.2312.11556} and VGBench~\cite{zou-etal-2024-vgbench} provide paired SVG and caption data to support tasks such as image generation from text and vice versa. While these resources enable research into SVG generation and representation learning, they are not designed for editing tasks where a transformation between two vector images must be described and executed.

Instruction-based image editing has seen rapid progress in the raster domain, with models like InstructPix2Pix~\cite{https://doi.org/10.48550/arxiv.2211.09800} enabling photorealistic editing based on natural language prompts. These approaches leverage large-scale datasets and diffusion models to modify pixel-based images.

Our work presents a new, challenging dimension of image editing by introducing a large-scale dataset for instruction-guided vector image editing, where the goal is to transform one SVG into another using a natural language instruction. Unlike prior work focused solely on generation or simple attribute edits, our dataset is structured around realistic transformations drawn from curated vector collections, supporting both fine-grained evaluation and more sophisticated model development.

While SVGEditBench~\cite{https://doi.org/10.48550/arxiv.2404.13710} explores instruction-based SVG editing, it focuses on six narrowly defined transformation types. In contrast, our dataset covers a much broader and more diverse set of edits, offering a more generalized benchmark for studying instruction-conditioned vector editing at scale.

Our work addresses this gap by introducing a dataset specifically for instruction-guided vector image editing, enabling new research in this underdeveloped area.

\section{Dataset}

Our dataset consists of 271,306 pairs of vector images, each accompanied by a natural language instruction that describes the transformation from the original image to its edited version. Each pair contains a source vector graphic, a target vector graphic, and a corresponding instruction detailing the editing operation required to achieve the transformation. This structure is designed to support training and evaluation of models for instruction-guided vector image editing. Example pairs can be seen in Figure~\ref{fig:example_pair}.

To facilitate robust evaluation, we split the dataset into three subsets:

\begin{itemize}
    \item Training set: 269,106 pairs
    \item Validation set: 200 pairs
    \item Test set: 2,000 pairs
\end{itemize}

To minimize data leakage, the splits were constructed by selecting entire individual collections. This approach reduces the likelihood that related vector styles or motifs appear across different subsets, supporting a more reliable evaluation of generalization to unseen styles and image structures.
   
This dataset is derived from SVG Repo~\cite{svgrepoRepoFree}, a large repository of free and open-source vector graphics that includes icons as well as various other high-quality vectors such as illustrations and design elements. Pairs of vector images were created by sampling within individual collections, which group graphics sharing a consistent visual style. Although images within a collection are not variations of the same design, their stylistic similarity provides a coherent context for generating natural language instructions that describe transformations between pairs.

SVG Repo serves as a proxy distributor of open-licensed vector content, indexing works from various public domain, open source, and user-submitted sources. Most content falls under the SVG Repo License, which permits sharing and adaptation without attribution, although credit is appreciated. In cases where specific licenses apply (e.g., MIT, GPL, CC BY), the corresponding terms are respected. All content was used in accordance with these licensing conditions.

In the following sections, we describe each phase in the dataset creation.

\subsection{Image Pair Sampling}

For an image editing task to be meaningful, the source and the target image should not only maintain overall style, but also share semantic components.

To sample sensible image pairs, we compared all images within each collection using CLIP-based~\cite{https://doi.org/10.48550/arxiv.2103.00020} image similarity on the rendered bitmaps. By computing pairwise similarity scores between images, we selected pairs with a similarity above a defined threshold, ensuring that the transformations are meaningful yet diverse. This process was applied only within the same collection to maintain stylistic consistency while capturing a range of visual changes suitable for instruction-based editing tasks.

In addition to CLIP, we also tested other similarity and clustering methods, including DINOv2~\cite{https://doi.org/10.48550/arxiv.2304.07193} visual embeddings, TF-IDF on image tags and generated captions, and pixel MSE between rasterized versions of the SVGs.

\begin{figure}[h]
    \includegraphics[width=\columnwidth]{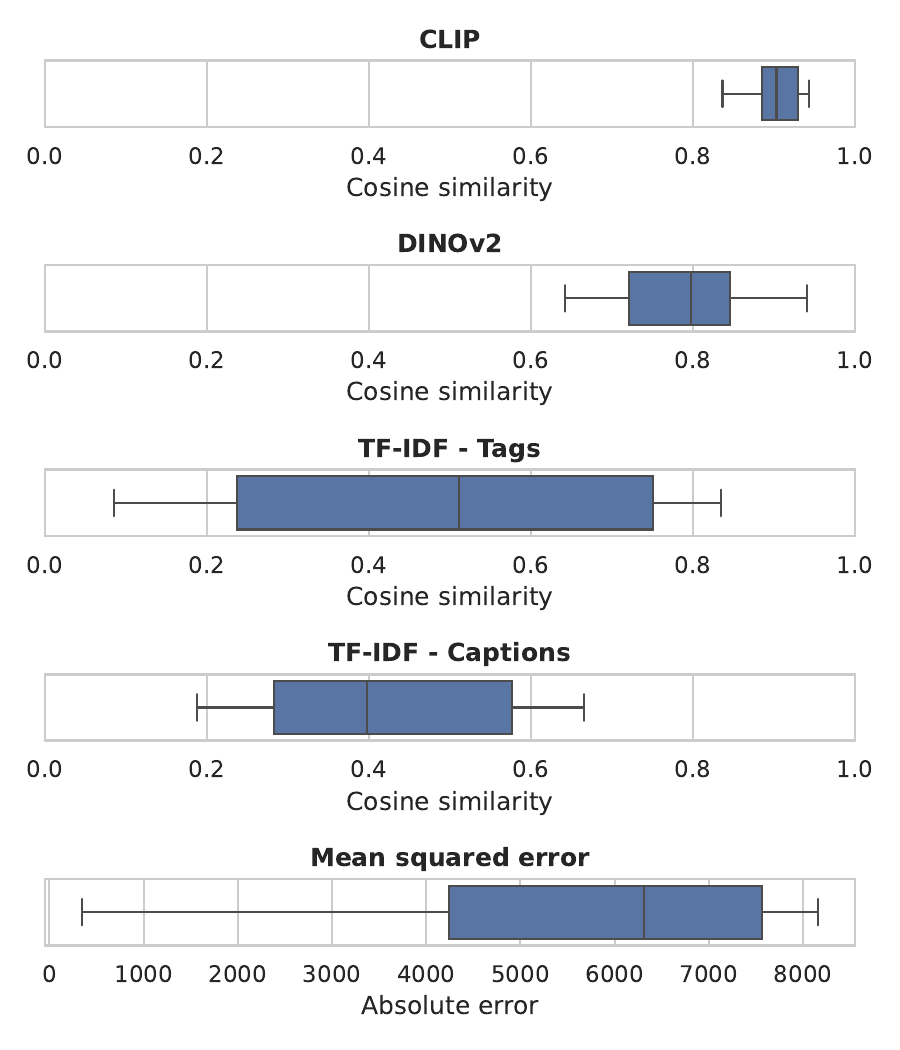}
    \caption{Distribution of manually selected similarity thresholds across 30 samples for five different pairing methods. Lower variance indicates more consistent and reliable pair selection.}
    \label{fig:tresholds}
\end{figure}

For each method, thresholds were tuned by manually going from most to least similar pairs in 30 separate collections to identify first pair that does not represent a meaningful editing transformations according to human judgment. The variability of threshold across collections for each method is illustrated in Figure~\ref{fig:tresholds}. CLIP was ultimately chosen due to its lowest threshold deviation, making it the most reliable and consistent approach.

\begin{figure}
    \includegraphics[width=\columnwidth]{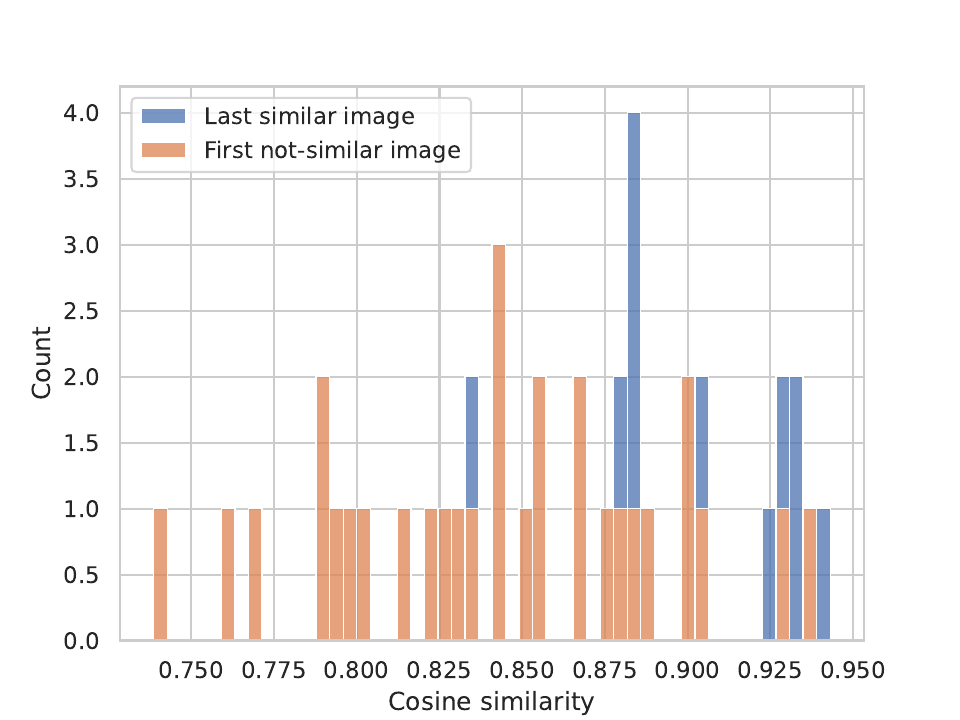}
    \caption{Histogram of CLIP similarity scores for the last image marked as similar and the first marked as not similar in a set of manually reviewed examples.}
    \label{fig:clip-histogram}
\end{figure}

Figure~\ref{fig:clip-histogram} documents how the manually annotated set of last relevant and first non-relevant pairs overlap in terms of chosen similarity measure. We select the threshold so that there are no false positives marked by the similarity on manually annotated set, i.e., above the most similar non-relevant pair.

After the similarity measure and threshold were established, we identified a total of 271,306 image pairs.

\subsection{Instruction Generation}

With the image pairs selected, the next step was to generate natural language instructions that describe the visual transformation from one image to the other.

Given the large scale of the dataset, edit instructions for individual pairs were generated automatically using vision-language models. We evaluated three different models for this task by manually ranking the quality of their generated instructions on a sample of 100 pairs.

Based on this evaluation, the models were ordered from best (1) to worst (3) in terms of instruction accuracy and relevance. The results of the ranking are shown in Table~\ref{tab:labeling_model_rank}.

\begin{table}[ht]
    \centering
    \begin{tabular}{l|c}
        \textbf{Model} & \textbf{Mean rank} \\
        \hline
        OpenAI GPT 4.1 & 1.28 \\
        \hline
        Llama 4 Maverick & 1.54 \\
        \hline
        Qwen2.5-VL 70B & 1.51 \\
    \end{tabular}
    \caption{Human ranking of labeling performance across three vision-language models. Lower mean rank indicates higher quality edit instructions, based on a manual evaluation of 100 image pairs.}
    \label{tab:labeling_model_rank}
\end{table}

Although Qwen2.5-VL 70B was not ranked as the best-performing model in our manual evaluation, it was ultimately selected for generating instructions across the full dataset. This decision was driven by practical considerations: Qwen2.5-VL 70B can be hosted locally, significantly reducing the cost and dependency associated with using API-based models like GPT-4.1 while supporting reproducibility and further development.

\section{Evaluating models}

\begin{table*}[!ht]
\centering
\begin{tabular}{l|c|c|c|c}
\textbf{Model} & \textbf{CLIP} ($\uparrow$) & \textbf{DINOv2} ($\uparrow$) & \textbf{MSE} ($\downarrow$) &
\textbf{Invalid count} ($\downarrow$) \\
\hline
\textit{Baseline -- no edit} & 0.9634 & 0.9011 & 10488 & 0 \\
\hline
\textit{Baseline -- white image} & 0.7188 & 0.2637 & 17262 & 0 \\
\hline
\hline
GPT-4o mini          & 0.9040 & 0.8058 & 8526 & 14 \\
\hline
Gemini 2.0 Flash & 0.9089 & 0.8135 & 11810 & 32 \\
\hline
Llama 4 Maverick & 0.9094 & 0.8331 & 9627 & 13 \\
\hline
Gemma 3 27B & 0.9105 & 0.8268 & 11881 & 66 \\
\hline
DeepSeek V3 0324 & 0.9203 & 0.8444 & 11194 & 278 \\ 
\end{tabular}
\caption{Comparison of CLIP, DINOv2 similarity, MSE distance, and invalid SVG counts across models. \textit{Baseline~--~no edit} uses the original unedited image as the output, while \textit{Baseline~--~white image} uses a blank white image. $\uparrow$~--~higher is better; $\downarrow$~--~lower is better}
\label{tab:model_comparison}
\end{table*}

Since instruction-guided vector image editing is a novel task with no specialized models available, we evaluate general pretrained large language models (LLMs) to establish a baseline performance on our dataset.
In our setup, the initial SVG image and the corresponding edit instruction are provided as input prompts to the model, which is then tasked with generating the edited SVG output.

\subsection{Metrics}

To evaluate model performance, we first rasterize both the generated and reference SVGs to 512×512 pixel images to enable visual comparison. We then use several metrics including Mean Squared Error (MSE), DINOv2 similarity, and CLIP score to compare the generated images against ground truth edited images. Following the findings of the Starvector~\cite{https://doi.org/10.48550/arxiv.2312.11556}, we emphasize that semantic similarity metrics like DINOv2 and CLIP better align with human judgments of output quality than purely pixel or geometry-based measures such as MSE. Additionally, we track the count of invalid SVG outputs, which are generated files that do not conform to SVG syntax or semantics, as a critical measure of model reliability. Thus, these metrics provide a more meaningful assessment of how well models understand and apply the editing instructions. 

\subsection{Results}
We evaluate model performance against two baselines to contextualize results. The first, \textit{Baseline~--~no edit}, uses the original, unedited SVG image as the model output, treating it as if the edit had been applied. The second, \textit{Baseline~--~white image}, uses a completely white image as the predicted edit. These baselines help illustrate the difficulty of the task and the strength of naive strategies.

\begin{figure}[t]
\centering
  \includegraphics[width=\columnwidth]{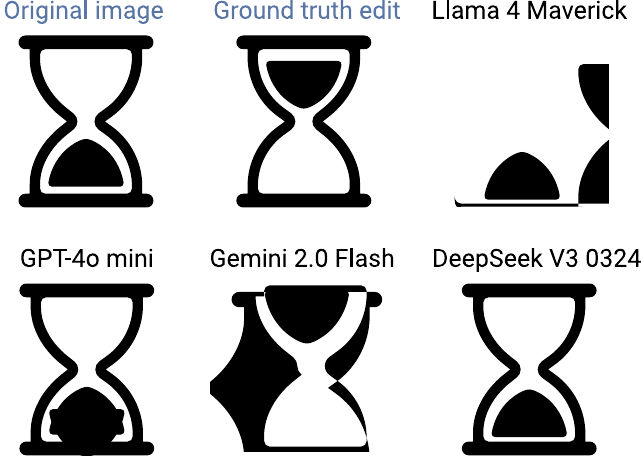}
 \caption{Failed edit attempts by several models on the hourglass example. The instruction was: \textbf{Move all the sand from the bottom half to the top half of the hourglass.}}
  \label{fig:edit_fails}
\end{figure}

All tested models performed significantly below expectations, failing to surpass the baseline strategy of leaving the original image unedited. Despite being prompted with both the initial SVG and a natural language instruction, models frequently produced incorrect or malformed SVG outputs. In contrast, the \textit{Baseline -- no edit}, where no changes were made to the input image, achieved higher similarity scores (CLIP, DINOv2) relative to the generated outputs. These results, summarized in Table~\ref{tab:model_comparison}, highlight the difficulty of the task and the current limitations of existing models in understanding and executing fine-grained vector editing instructions. A representative example of failed attempts is illustrated in Figure~\ref{fig:edit_fails}, where several models struggle to correctly execute a simple sand movement instruction on an hourglass image.

\section{Conclusions}
We introduced a large-scale dataset for instruction-guided vector image editing, filling a gap in existing research focused primarily on raster images. Our dataset enables the development and evaluation of models that interpret natural language instructions to perform structured edits on vector graphics. Initial experiments with state-of-the-art models show that this task remains highly challenging, with current approaches failing to outperform a simple no-edit baseline. We hope that our dataset will encourage and support further research into language-driven vector manipulation and the development of models better suited to structured visual domains.

\section*{Limitations}

Although our dataset provides a valuable resource for instruction-guided vector editing, it has some limitations. The clustering approach based on similarity thresholds, although effective, may introduce errors and may occasionally pair images with less meaningful or ambiguous transformations, despite the fact that our methodology sets the thresholds such that the risk of these errors is minimized. Additionally, the automatic generation of edit instructions using vision-language models may occasionally introduce noise and inconsistencies, as these models are not perfect in understanding all visual differences. Another limitation is that some instructions may contain excessive detail, effectively allowing a model to generate the "edited" SVG from scratch without relying on the source image. Future work could focus on refining clustering methods and improving instruction quality through human-in-the-loop verification or more advanced labeling techniques.

\bibliography{custom}

\appendix

\end{document}